\documentclass[journal]{IEEEtran}
\usepackage{cite}
\usepackage[T1]{fontenc}
\usepackage{booktabs}
\usepackage{amsmath,amsfonts}
\usepackage{multirow}
\usepackage{makecell}

\usepackage[center]{caption}
\ifCLASSINFOpdf
  \usepackage[pdftex]{graphicx}
  
\else
\fi

\begin{document}
\bstctlcite{IEEEexample:BSTcontrol}

\title{Fragment driven dual-stream CNN Network with attention for Offline Writer Identification using Word Data}

\author{
    Vineet~Kumar\thanks{Vineet Kumar and Suresh Sundaram are with the Department of Electronics and Electrical Engineering, Indian Institute of Technology Guwahati, Guwahati 781039, India (email: vineet18@iitg.ac.in, sureshsundaram@iitg.ac.in).}, 
    Suresh~Sundaram
}
\maketitle

\begin{abstract}
 
In this paper, we  propose a text-independent writer identification system utilizing a   dual-stream Convolutional Neural Network (CNN) trained  solely using   fragments extracted from word images.  More specifically,  we employ the integration of feature maps from the classification and dissimilarity frameworks   to extract     features  specific to a writer as well as
cues, that are prevalent across fragments of several writers.  This aspect makes our proposal different to previous works on convolution networks, that are trained on writer-specific word images and their corresponding fragments. Furthermore, we explore an attention mechanism that refines the ability to discern relevant features by emphasizing important regions within the fragments.  Experiments conducted on word images from the IAM, CVL, and CERUG-EN databases suggest that the proposed network architecture captures well the intricate characteristics of fragments contributed by different writers. The results obtained are promising to previous end-to-end  deep learning frameworks proposed in the literature. 

\end{abstract}
\IEEEpeerreviewmaketitle
\section{Introduction}
Handwriting is a form of behavioural biometric \cite{Chi:2018} having a distinguishing trait and has been effectively employed in fields such as forensic analysis \cite{Fernandez:2010}, historical document analysis \cite{Bulacu:2007,He:2014}, and security \cite{Faundez:2020} to name a few. Writer identification involves presenting a list of likely writers in order to identify the authorship of a document from a collection of reference documents stored in the database. Based on the methodology employed to acquire the   data it can be broadly grouped into offline and online \cite{Rehman:2019}. In the former, the temporal attributes such as coordinates, pressure, and angle is recorded by specific devices such as a tablet. As contrast to it, in the latter, the document containing handwritten text is   captured in the form of an image and analyzed subsequently. Based on the content of the handwritten text, writer identification systems are   categorized into text-dependent and text-independent approaches. In the text-dependent approach, the content of the handwritten text is fixed, while, in the text-independent approach, the content of the handwritten text can vary. \\  \indent Traditional methods for writer identification rely on extracting handcrafted features to analyze and capture the unique characteristics of a writer's style from handwritten documents. These features are specifically designed to highlight attributes such as curvature \cite{Bulacu1:2007}, slant \cite{Siddiqi:2010}, and ink width \cite{Matri:2001}, among others. Since these handcrafted features depend on utilizing statistical information from handwriting, they typically require a significant amount of textual data, such as a paragraph or a block of text containing multiple sentences \cite{Bulacu:2007, Khan:2019} to achieve reliable performance.  \\  \indent 
Writer identification systems utilizing handcrafted features achieve high recognition rates on public datasets. However, there may be   applications for which access to handwritten data is limited. In such cases, decisions may need to be made based on a very small amount of handwriting, such as individual words. This scenario presents a significant challenge, as the writer-specific style information is reduced compared to page or text-level inputs. Consequently, handcrafted feature-based systems fail in these situations, yielding low recognition rates. \\  \indent
The pioneering exploration  in the   direction of word-level based off-line writer identification  is that of  \cite{He:2019}. Here,  the authors employ a multi-task framework to enhance writer-related information by incorporating attributes learned in the auxiliary task along with   features from the main task. In a subsequent work \cite{He:2020}, the same authors proposed a deep neural network (\textit{FragNet}) to extract powerful features from  the input word images. This network consists of two pathways: the feature pyramid pathway  and fragment pathway. The former is used for feature map extraction  while the latter is trained to identify the writer using fragments extracted from the input image in combination with the feature maps produced by the feature pyramid. \\  \indent
In  another contribution \cite{He:2021}, the authors introduce an end-to-end neural network system known as global-context residual recurrent neural network  (GR-RNN) for identifying writers from handwritten word images. The system combines global context information and a sequence of local fragment-based features. Further, in order to capture the spatial relationships between these fragments, a recurrent neural network is used to enhance the discriminative power of the local fragment features. \\  \indent Moving ahead,  in \cite{Zhang:2022},   a Residual Swin Transformer classifier   is designed to capture both local and global handwriting styles effectively from single-word images. The model employs transformer blocks to handle  the local information by interacting with individual strokes and utilizes holistic encoding with the identity branch and global block to capture   handwriting characteristics. In a work published in 2022, the authors of \cite{Srivastava:2022} proposed a Spatial Attention Network to identify writers based on the notion that specific regions of word images have  shape information
 unique to a writer. Along with this, a Multi-Scale Residual Fusion Classification network and Patch net framework (similar to the idea of \emph{Fragnet}) was also investigated.

\section{Research framework}

The aforementioned prior  works considered networks that were trained  with the handwritten word as the input, (that provides the global information) along with the sequence of its constituent fragments (capturing the local characteristics).  In our proposal, we deviate from this strategy  by introducing   a dual stream convolution-based framework whose parameters are trained {solely} by employing the  fragments of the word images.  The utility of fragments as input to the network ensures that these handwritten ink traces, characteristic of the writer are processed irrespective of their positions in the   word \footnote{ Typically, fragments extracted from a word comprise of characters and parts of characters.}.   As such, the scores obtained from the fragments of a word are summed to determine the identity of the writer.    We demonstrate in  Section \ref{compar}  that our proposal   is effective in providing results that are competing with other state of art  end to end deep learning methods that are trained on both handwritten words and their fragments.  At this point, it is worth taking cognizance of our previous works   \cite{KUMAR1,KUMAR2}, that are   based on analyzing the fragments of a word using a LeNET and Siamese architecture. However, those are {not} end-to end as the extracted deep learning features from the fragments constituting the word are processed  by a SVM classifier after incorporating the idea of saliency.   \\ \indent Convolutional networks that rely solely on classification-based frameworks typically perform well when distinguishing between  dissimilar samples. However, for fine-grained tasks such as writer identification, these approaches often struggle to capture the subtle variations between similar handwriting styles—especially when working with short word images that contain limited textual content. Keeping this in perspective, we introduce a dual stream approach  that integrates classification and dissimilarity learning frameworks on the word fragments to achieve more discriminative feature representations. While the classification branch captures localized, writer-specific features from the fragments, the dissimilarity  branch focuses on learning pairwise similarities, enabling the model to distinguish fine-grained variations that lie beyond predefined class boundaries. For this, we consider utilizing a  Siamese architecture that is first pre-trained on a  writer independent   dataset but then subsequently fine-tuned using the fragments of the words.  An integrated attention mechanism further enhances our framework by highlighting the most informative regions within each fragment. This attention-guided,  learning strategy on the fragments significantly improves the ability of the model to identify subtle  cues, leading to better performance in writer identification tasks. \\ \indent In summary, the main contribution of this paper include: \\
  1) We suggest  a proposal that  integrates classification and dissimilarity frameworks in a dual channel network to extract both localized, writer-specific features and  writer-independent features (that go beyond predefined class boundaries).  More specifically, as we shall see in the sub-section \ref{dual}, the writer-dependent module focuses on capturing fine-grained characteristics specific to an individual writer at the fragment level. In contrast, the writer-independent module aims to learn representations of fragments by leveraging the shared characteristics contributed by all writers enrolled in the system.
\\
2)  The dual-stream network is enhanced with an attention mechanism that highlights important regions within the fragments, improving feature recognition and boosting performance in writer identification tasks. \\
  3) Comprehensive experiments conducted on three databases show that the proposed method outperforms existing approaches, highlighting its effectiveness . \\  \indent Owing to the fact that our approach is text independent and processes fragments   irrespective of their positions in the
word,  we do {better}  compared to the Swin transformer (with position encoding) proposed in \cite{Zhang:2022} 
(refer Table \ref{prior_wrk}).  We opine that the  transformer, though effective in  providing attention ignores the fact that the fragment patches can appear anywhere in a word. Position encoding suggests that the fragments have to necessarily appear in a given order - an aspect that is not feasible in text independent writer identification systems.

\section{Proposed Method}\label{sec3}
In this section, we present the details of the proposed method for writer identification. Our strategy leverages a framework based on Convolutional Neural Networks and attention mechanism. We process image fragments extracted from word images through two parallel CNN modules: one trained in a writer-dependent manner to capture writer-specific details from the fragments and the other employing a writer-independent approach to extract features of fragments that are prevalent across several writers. The combined features (post attention) are passed through a classification block with fully connected layers, which assign probability scores to the fragments using a softmax activation function. The scores are then accumulated across all fragments of a word image, and the  identity is determined based on the maximum accumulated score.  Further, the attention mechanism  leverages relationships within the convolution layer feature maps, highlighting critical regions in the fragments. By generating attention weights, the network refines its ability to discern relevant features, thus improving performance in writer identification.  



\subsection{ Proposed Dual-stream CNN Model} \label{dual}

\begin{figure*}[!ht] 
\centering
\includegraphics[width= \textwidth]{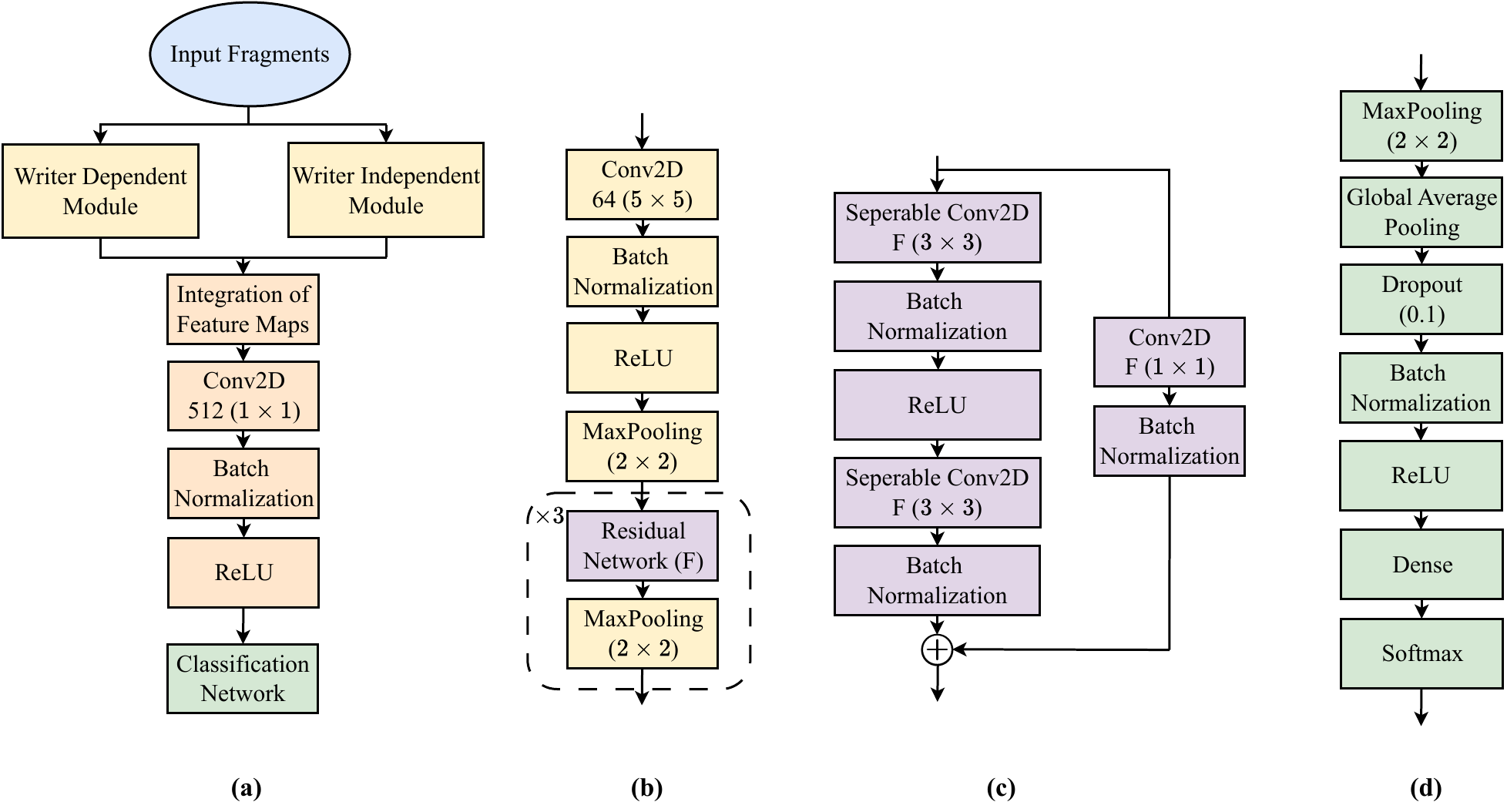}
\caption{The architecture of dual-stream CNN model used for training the fragments of the words. Each convolution block and its variant (represented as  Conv2D/Separable Conv2D) is followed by entries signifying the number of filters and their kernel size, respectively. The sub-figure (a) represents the overall architecture of the proposed dual stream network. Sub-figure (b) depicts the architecture of the writer-dependent module and the  writer independent module. However, for pre-training of the writer dependent module in a Siamese framework, we consider augmenting  the blocks of   global  average pooling, dropout, and Fully Connected
layers after the  last residual network. Sub-figure (c) represents the structure of the Residual block used in the writer dependent and independent modules, and while (d) presents the architecture of the classification block.}
\label{cnn_model}
\end{figure*}

 To begin with, we obtain the  fragments    by dividing the word image into  $P \times P$ patches of same size.  Subsequent to this, each of the fragments need to be mapped to a feature vector for subsequent analysis. For this, we propose a multi stream  Convolution Neural Network (CNN) of the form shown in Figure \ref{cnn_model} (a).  \\  \indent The network comprises two parallel modules: the writer-dependent   and the writer-independent module. The writer-dependent module is designed to capture fine-grained local handwriting characteristics of the fragments, whereas the writer-independent module focuses on modeling broader handwriting patterns prevalent across several users. To enhance the representation power of the system, we aim to integrate the feature map representations from the writer-independent module with those  from the writer-dependent module. 

\subsubsection{Writer dependent (\textit{WD}) module}
Since each writer exhibits a unique handwriting style, the strokes within handwritten words tend to display strong internal consistency. Consequently, understanding the relationships between different strokes is essential for effectively modeling local handwriting characteristics. The purpose of the writer-dependent Convolutional Neural Network (CNN) module is to extract writer-specific features from the input fragments.

As illustrated in Figure \ref{cnn_model} (b), the architecture of the writer-dependent module comprise convolution layers with residual connections,  followed by Batch Normalization and a Rectified Linear Unit (ReLU) activation function. The convolution layers  in the residual network employ filters of 128, 256 and 512 channels, with kernel sizes of ($n \times n$), where $n \in {1, 3}$ (refer Figure \ref{cnn_model} (c)). The first convolution layer  uses a ($5 \times 5$) kernel to capture broader structural patterns in the fragment image.The skip connections in the residual block facilitate the network  to learn complex features while alleviating the vanishing gradient problem.
\subsubsection{Writer independent (\textit{WI}) module}\label{WI_module}
The writer-independent module is designed to extract features from input fragments,  that tend to remain consistent across different writing instances of the  individuals enrolled in the system. These patterns can be incorporated into writer identification, particularly in unconstrained settings.

Traditional classification-based frameworks are well-suited for identifying fine-grained local features within predefined class boundaries. However, they often fall short in modeling relationships among similar samples across writer categories, as their primary objective is label assignment rather than measuring inter-sample similarity. To overcome this limitation, we adopt a dissimilarity-based framework that emphasizes pairwise comparisons. This approach facilitates the learning of stylistic consistencies by evaluating the degree of  dissimilarity between pairs of  fragment samples. The dissimilarity measure employed is a distance metric in a learned embedding space, wherein fragments with stylistic similarities are mapped closer together. For this purpose,  we employ a Siamese architecture pre-trained using the triplet loss function.

The Siamese network is pretrained on the Omniglot dataset \cite{Omniglott, Koch:2015} , which was chosen for its diverse range of character styles and its suitability for one-shot image recognition tasks \footnote{In order to support the claim on the diversity range of character styles of the Omniglot dataset, we compare its performance in sub-section \ref{indper} with the EMNIST dataset.}.  Architecturally, each of  the CNN pathways of the Siamese  writer-independent network  mirrors the writer-dependent model (presented in Figure~\ref{cnn_model}(b)), with the addition of global  average pooling, dropout, and fully connected layer after the third residual block.

The pre-training of the writer independent block is conducted using the TensorFlow framework, with the network optimized via the Adam optimizer \cite{kingma:2014} over 100 epochs. The learning rate is initialized at 0.001, and a weight decay factor of 0.5 is applied when the validation loss fails to improve over five consecutive epochs. Upon integration with the \textit{WD} module, we leverage the idea of transfer learning by fine tuning some of the later layers  with the fragments of the word.
\subsubsection{Classification block}
The feature representations obtained from the writer-dependent and writer-independent modules are fused to form a combined feature $(\hat{F})$ mathematically represented as:
\begin{equation}
    \hat{F}=f_{c}(F_{\textit{WD}},F_{WI})
\end{equation}
Here $F_{WI}$ and $F_{\textit{WD}}$ represents the writer-dependent and independent features, respectively, while $f_c$ denotes the combining function.  The combined feature $(\hat{F})$  is obtained by applying the combining function $f_c$ to the feature maps extracted from the outputs of the third residual block of both branches.  In our study, we have explored different methodologies to obtain an efficient combining function, the details of which is discussed in section \ref{sec7}-C.  \\  \indent
The combined feature maps are   then passed to the classification block, which consists of a global  average pooling (GAP) layer, a dropout layer, and a Fully Connected (FC) layer.  The detailed architecture of the classification block is illustrated in Figure \ref{cnn_model}(d). 

The   network in Figure \ref{cnn_model} (a) is trained in an end-to-end framework, where the weights of the writer-dependent module are updated from scratch, while the writer-independent module leverages transfer learning. Specifically, the weights up to the second residual block of this module are kept frozen, and fine-tuning is performed on the subsequent layers to adapt the pre-trained parameters to the fragments of the input word.

\subsection{Attention module} \label{attn}
\begin{figure}[t] 
\centering
\includegraphics[width=.48\textwidth] {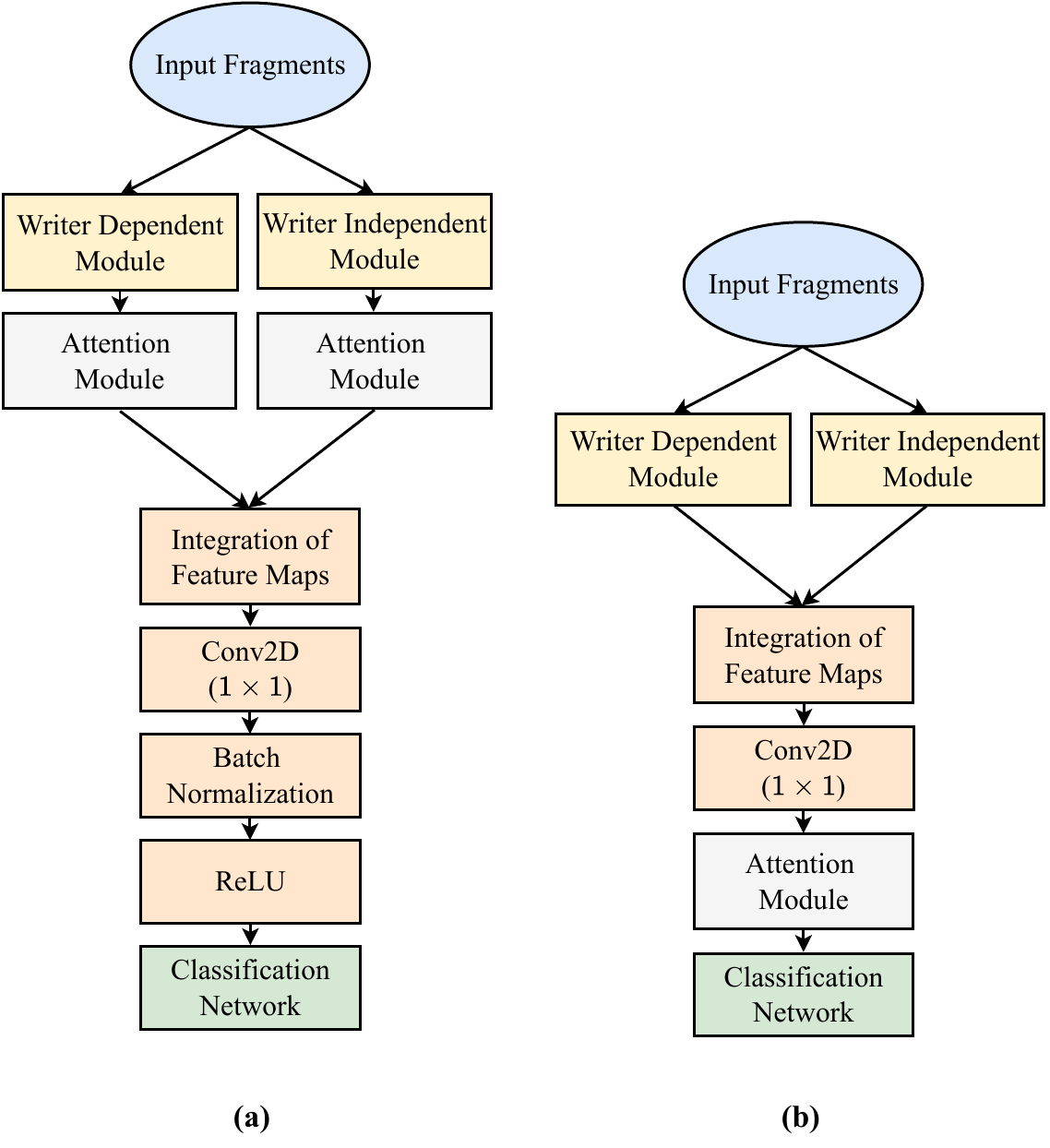} 
\caption{Block diagram representing different configurations of attention module. In Figure (a) the attention module is incorporated separately to the writer dependent and independent modules. In Figure (b), the attention module is  applied after the integration of the  feature maps. }
\label{attention_config}
\end{figure}

In this work, we propose  an attention mechanism to selectively focus on specific elements within the fragments of a word, thus assigning varying levels of importance to different regions. This adaptability boosts the capability of the model to discern complex patterns and relationships, thus enabling the combined network  of Figure \ref{cnn_model} (a) to learn more informative features. \\  \indent
In our research, we have explored two configurations of the attention module. In the first configuration, the attention module is  applied separately to both the writer-dependent and writer-independent modules before their features are combined.  Contrast to this, in the second configuration,  the attention module is considered post the integration of features of the writer-dependent and writer-independent modules.  These configurations are depicted in Figure \ref{attention_config} (a) and (b) respectively. The evaluation of these configurations in the writer identification system is detailed in Section \ref{sec7}. \\  \indent
Our approach utilizes a MobileViT-inspired multi-head attention mechanism,  outlined in \cite{Mehta:2021}. The input feature map $\mathbf{X} \in \mathbb{R}^{H \times W \times C}$ is reshaped into a two-dimensional matrix $\mathbf{X}_U \in \mathbb{R}^{N \times C}$, where $N=H\cdot W$ denote the spatial dimensions of ${X}$. 
An encoder block $G$ projects $\mathbf{X}_U$ into a $d$-dimensional embedding space.
\begin{equation}
    Z = G(X_U) \in \mathbb{R}^{N\times d}
\end{equation}
This forms the input to the multi-head self-attention module. For each attention head \(i \in \{1, \dots, h\}\), three separate trainable projection matrices are used  namely $W_i^Q,W_i^K$, and $W_i^V$ each of size $d\times d_h$, where $d_h=d/h$ and $h$ represents the number of heads. These projection matrices transform $Z$ into Query $(Q_i)$, Key ($K_i$), and Value ($V_i$) as:
\begin{equation}
Q_i = Z W_i^Q,\quad K_i = Z W_i^K,\quad V_i = Z W_i^V,
\end{equation}
 Here, $Q_i$ represents what each token is looking for, $K_i$ represents what each token contains, and $V_i$ represents the actual information passed on. The attention weights are computed as:
\begin{equation}
A_i = \mathrm{softmax}\left(\frac{Q_iK_i^\top}{\sqrt{d_h}}\right),
\end{equation}
and the output of $i^{th}$ head is given by
\begin{equation}
\mathrm{head}_i = A_i V_i.
\end{equation}
The outputs from all heads are concatenated and projected back to $d$ dimension:
\begin{equation}
X_t = \left[\mathrm{head}_1,\dots,\mathrm{head}_h\right] W^O,
\end{equation}
where \(W^O \in \mathbb{R}^{d\times d}\) is the output projection matrix.

The resulting representation \(X_t\) is passed through a decoder block \(\hat{G}\) to map the features back to the original \(C\)-dimensional space:
\begin{equation}
\widehat{X}_U = \hat{G}(X_t) \in \mathbb{R}^{N\times C}.
\end{equation}

\begin{figure*}[!ht] 
\centering
\includegraphics[width=.85\textwidth] {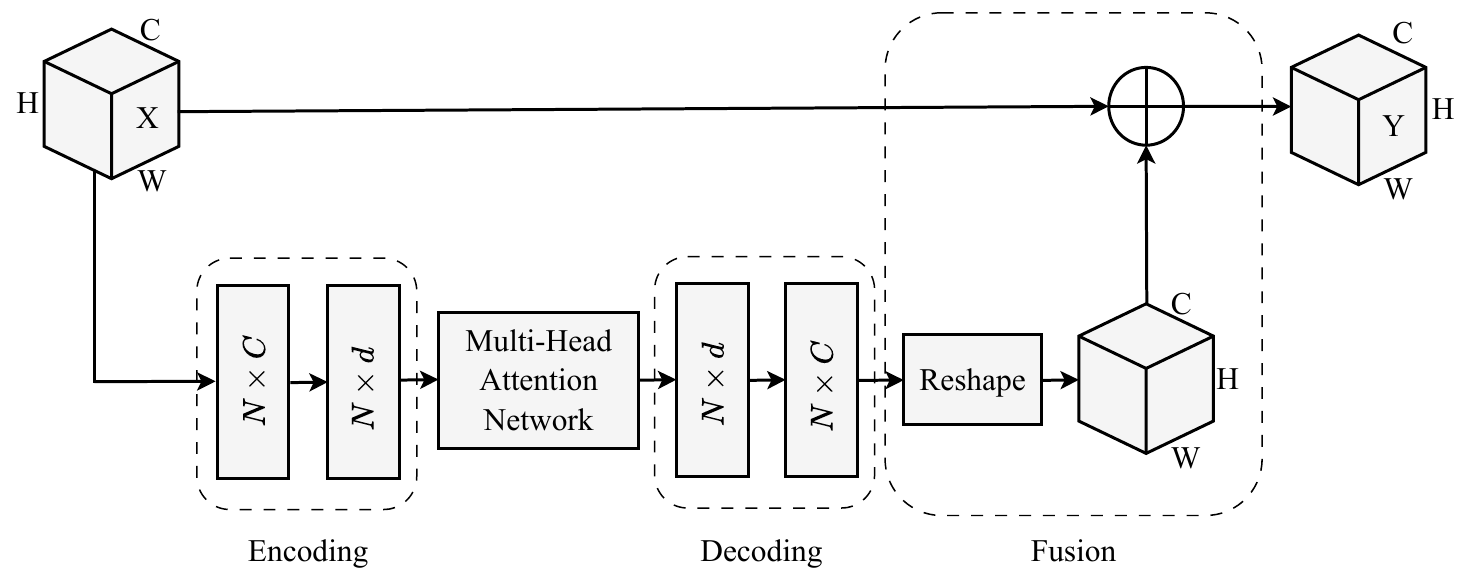} 
\caption{Block diagram representing configurations of the attention module.}
\label{attention_module}
\end{figure*}

After reshaping the resulting decoder output to \(\mathbb{R}^{H\times W\times C}\), it is added to the original input feature map \(X\), to fuse the local and global contextual representations. Figure \ref{attention_module} shows the pictorial overview of the attention module.




\subsection{Implementation details}\label{sec6}
The fragments extracted from the word image serve as input to the unified model, which integrates both the writer-dependent and writer-independent modules. During training, gradient backpropagation is guided by the label smoothing cross-entropy loss function \cite{szegedy2016}, which converts the one-hot encoded hard labels $y_i$ into soft labels $\hat{y}_i$ as follows:

\begin{equation}\label{lbs}
\hat{y}_i=y_{i}(1-\epsilon)+(1-y_i)\frac{\epsilon}{K}=
\begin{cases}
1-\epsilon,& {i=target} \\
\frac{\epsilon}{K}, & {i \neq target}
\end{cases}
\end{equation}

Here $K$ and $\epsilon$ denote the number of writers and the label smoothing factor respectively. Using this approach, the total loss is obtained by computing the cross-entropy loss between the predicted score in the range $(0,1)$ with the adjusted ground truth label.

\begin{equation}
L_i=-\sum_j^W \hat{y}_{ij}. \mbox{log}(p_{ij})
\end{equation}
where  $\hat{y}_{ij}$ and $p_{ij}$ denote the modified ground truth and the predicted score, respectively, for the $i^{th}$ fragment of the $j^{th}$ writer. The overall training loss is subsequently computed by averaging across  the  $N_{T}$  word fragments from the writers enrolled in the system.
\begin{equation}
L=\frac{\sum_i^{N_{T}} L_i}{N_{T}}
\end{equation}

In the testing phase, given a word image $w$ yielding $N$ fragments, the   overall score  $P_i(w)$    associated with the $i^{th}$ writer is computed by averaging the responses corresponding to all the fragments as

\begin{equation}\label{averaging}
P_i(w)=\frac{\sum _{k=1}^{N }p_{ki}}{N}
\end{equation}
In the above Equation,  $p_{ki}$ denotes the individual   score of the $k^{th}$ fragment corresponding to the $i^{th}$ writer. The  identity of the word image $w$ is assigned to the  writer with the highest score.

\begin{equation}\label{classification}
\phi(w)=\underset{i\in \{1,...,W\}}{argmax}\; P_{i}(w)
\end{equation}

Our proposed network is implemented on TensorFlow framework. The batch size is set to 16 for training. The network is optimized using  the Adam optimizer \cite{kingma:2014} with an initial learning rate of 0.001. The weight parameters are decayed by a factor of 0.5 whenever  the validation accuracy has stopped improving for 10 continuous epochs. The number of epochs for training the network is set to 150. The label smoothing factor $\epsilon$ in Equation \ref{lbs} is set to 0.1.

Each word is divided into fragments of equal size.
With regards to each of the fragments, they are   resized to  $105 \times 105$ while maintaining the aspect ratio and padding (if necessary) with white pixels. These modified fragments are then passed
through the dual-stream CNN module to obtain scores that are then accumulated to get the identity of the writer. In our implementation, we use nine fragments .
\section{Experiments}\label{sec7}
\subsection{Dataset Description}
The proposed method is evaluated  on the IAM \cite{IAM}, CVL \cite{CVL} and CERUG-EN \cite{Schomaker:2015}. The IAM database contains handwriting samples collected from 657 writers. Each writer has contributed a variable number of handwritten documents, 301 writers have contributed more than two handwritten documents while the rest have contributed only one handwritten document. In our implementation the IAM dataset is modified as described in \cite{Wu:2014}. For writers contributing more than one page of a handwritten document, we randomly select two pages, of which one page is used for training and the other for testing. For writers with only only one page of a handwritten document, we split it roughly into two halves. One half is used for training and the other half for testing. We utilize the bounding box information for the used images provided in the dataset to generate the training and test samples.\\  \indent The CVL dataset contains handwritten documents collected from 310 writers of which 27 writers have contributed 7 documents (6 in English and 1 in German) and the rest 283 have contributed 5 documents (4 in English and 1 in German). In our experiment, we utilized only handwritten English documents. We follow the same methodology as is done in \cite{He:2020}, by selecting three documents per writer for training and the rest pages for testing. Similar to the IAM dataset, segmented word images are made available in this dataset \\  \indent  The CERUG-EN dataset is a collection of handwritten documents from 105 subjects, with two English paragraphs contributed by each subject. In our experiment, the first paragraph written by each subject is used for testing and the second paragraph is used for training. The authors of \cite{Schomaker:2015} make the segmented word images accessible in the database, just like the IAM and CVL databases.

\begin{figure}[t] 
 \centering
 \includegraphics[width=.45\textwidth]{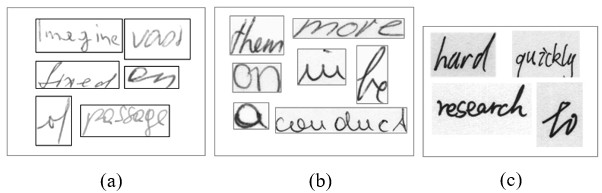}
\caption{Examples of training samples in (a) CVL database, (b) IAM database, and (c) CERUG-EN database. }
 \label{fig:db}
\end{figure}

The segmented word images of enrolled writers are included in IAM and CVL databases. However, in the case of CERUG-EN database, the segmentation was done manually by the authors of \cite{He:2020}. Table \ref{datset} gives a detailed overview of the datasets used in our experiments, and Figure \ref{fig:db} shows some of the training samples of the dataset. 
 
\begin{table}[t]
\centering
\caption{Overview of the datasets used in experiments with the number of training and testing word image samples. }
\label{datset}
\begin{tabular*}{.45\textwidth}{@{\extracolsep\fill}cccc@{}}
\toprule
Dataset   & Number of writers & Training words & Testing words \\ \midrule
IAM       & 657               &  56432 &  25827  \\
CVL       & 310                &  62406 & 34564  \\ 
CERUG-EN  &105                 & 5702 &  5127  \\\bottomrule
\end{tabular*}
\end{table}

In the following subsections, we present an ablation study on each of the modules in the proposed system. The contribution of individual  modules is analyzed through a series of experiments, highlighting their impact on the overall performance.
\subsection{Performance evaluation of the writer-dependent (WD) and writer independent (WI) module} \label{indper}
 In the first experiment, we compare the individual performance of the baseline writer-dependent (\textit{WD}) module with that of the  writer-independent (\textit{WI}) module. The writer-dependent module is trained end-to-end using   fragments extracted from the words of a specific writer. In order to obtain the identity of the user from this module alone, we consider the end to end  version of Figure \ref{cnn_model} (b) by  augmenting the  blocks of   global  average pooling, dropout, and  a fully connected
layer after the  last residual network. \\ \indent
 In contrast, the writer-independent module leverages the transfer learning idea  wherein the layers of the CNN are first pre-trained using a Siamese architecture with a triplet loss function.  It may be worth reminding here that the network being implemented is the end to end version of Figure \ref{cnn_model}  (b) obtained by  augmenting the  blocks of   global  average pooling, dropout, and  a fully connected
layer after the  last residual network.  Subsequent to pre-training, writer fragments are passed through one branch of the Siamese network and the parameters after the second residual network are fine tuned for evaluation in a multi-label  classifier setting \footnote{In this setting, we consider the output layer size to correspond to the number of writers enrolled in the system.}    using the label smoothing cross entropy loss.  In this ablation study, we consider the  performance of the  fine-tuned \textit{WI} modules that have been  pre-trained on two separate datasets: EMNIST \cite{Cohen:2017} and Omniglot \cite{Omniglott}.  
  The primary objective of using two different datasets for pre-training the writer-independent network is to assess the influence of dataset choice on its performance.  \\  \indent  Table \ref{Writer Dependent} presents the performance of the baseline writer-dependent module, while Table \ref{Writer Independent} shows the results of the writer-independent module pre-trained first on the EMNIST and Omniglot datasets and subsequently fine tuned. The evaluation of the module is conducted across different dimensions of the embedded feature representations.  \\  \indent
Based on the results presented in Tables \ref{Writer Dependent} and \ref{Writer Independent}, it can be observed that the baseline  \textit{WD} module trained directly on writer fragments outperforms the network trained on writer-independent datasets. Moreover, from Table \ref{Writer Independent} following observations can be made: 
 \begin{enumerate}
     \item  The performance of the \textit{WI} module initially improves as the size of the embedding layer increases with the best accuracy obtained at 512 dimensions. Furthermore, it can be inferred  that  higher embedding dimensions (such as 1024 and 2048)  introduce redundant information that does not contribute meaningfully to the task,  resulting in a  drop in performance. 
     \item The \textit{WI} module pre-trained on the Omniglot dataset outperforms  to the one  on  the EMNIST. This is primarily owing to the  handwriting samples of Omniglot that encompass  varied character styles, promoting the learning of diverse writer-independent features. This diversity reduces the risk of the network overfitting to dataset-specific characteristics, thereby enhancing its ability to learn better across different writers  with varying styles.
      \end{enumerate}
In the light of the above aspects, hereinafter, for the rest of the experiments, we consider the   pre-trained parameters of the \textit{WI} module corresponding to an embedding dimension of 512, as obtained from the Omniglot database. 
\begin{table}[t]
 \centering
 \caption{Comparison of average identification rate (in \%) for the baseline writer-dependent (\emph{\textit{WD}}) module trained on word image fragments.}
 \label{Writer Dependent}
 \begin{tabular}{ccc}
 \toprule
 {Database} &Top 1 & Top 5 \\ \midrule
{CVL}      & 89.76         & 95.46         \\
 {IAM}      & 92.07        & 96.43       \\
 {CERUG-EN} & 96.05         & 99.56         \\ \bottomrule
 \end{tabular}
 \end{table}
\begin{table}[t]
\caption{Average writer identification rate (in \%) based
on the word image fragment representations obtained from
the penultimate layer of the \textit{WI} module. The
pre-training of the module is done on the samples of the
EMNIST and Omniglot datasets with varying sizes of the penultimate layer.
The best identification rate is marked in bold.}
\label{Writer Independent}
\resizebox{.48\textwidth}{!}{%
\begin{tabular}{ccccllllc}
\hline
\multirow{3}{*}{Testing Database} & \multicolumn{8}{c}{Training Database}                                                                                                 \\ \cline{2-9} 
                                           & \multicolumn{4}{c|}{EMNIST}                                             & \multicolumn{4}{c}{Omniglot}                      \\ \cline{2-9} 
                                           & {256} & {512} & {1024} & \multicolumn{1}{l|}{{2048}} & {256} & {512} & {1024} & {2048} \\ \hline
{CVL}                               & 80.44       &{82.01}         & 81.32         & \multicolumn{1}{l|}{81.49}         & 83.47       & 87.43        & {86.70}          & 85.58          \\
{IAM}                               & 64.52        &  {69.18}        & 69.04         & \multicolumn{1}{l|}{68.24}         & 76.39        & {81.43}        & 79.61         & 78.84         \\
{CERUG-EN}                          & 59.56        & 69.74        &66.06          & \multicolumn{1}{l|}{{63.55}}         & 69.14        & {73.28}            & 70.56         & 67.39         \\ \hline
\end{tabular}%
}
\end{table}
\subsection{Impact of the dual stream network without attention}
In the next experiment, we study the performance of the dual stream network obtained by combining the writer-dependent and independent modules. We explore three distinct strategies for fusing writer-dependent and writer-independent features. In the first configuration labelled as \emph{WI+\textit{WD}+Max} , the maximum of the feature maps produced by the last residual layers of the two branches is obtained.  In the second strategy (\emph{WI+\textit{WD}+Add}), the feature maps produced by the last residual layers of the two branches are combined using pixel-wise addition while  in the third (\emph{WI+\textit{WD}+Concat}), the feature maps from both branches are concatenated and passed through a convolutional layer. The resulting fused features then serve as the basis for the classification module. 

Table \ref{ablation study1} demonstrates that integrating the writer-independent (\textit{WI}) module with the baseline writer-dependent (\emph{\textit{WD}}) model leads to a notable improvement in overall system performance. This enhancement can be attributed to the complementary strengths of both modules. The writer-dependent component is adept at capturing fine-grained, writer-specific local features by training directly on fragments from the target writers. However, its ability to learn diverse patterns is often limited, and constrained by the amount of text available for training.
However upon integration with  the writer-independent module,   the  resulting combined model is able  to capture cues that encompass variations in writing style and character morphology.

Furthermore, the use of concatenation followed by convolution for combining dual-stream features proves more effective than the other two methods of combining the write-dependent and independent features. Concatenation preserves the full set of features from each stream without enforcing assumptions about their spatial or semantic alignment. This enables the subsequent convolutional layer to learn an optimal fusion strategy, allowing for richer feature interactions and greater representational flexibility. In contrast, the other two strategies merge feature maps in a fixed, non-learnable manner, assuming alignment and similarity that may not exist, particularly when the streams encode distinct or heterogeneous information.

\begin{table}[t]

\caption{Comparison of average identification rate (in \%)  using different configuration of fusing writer dependent (\textit{WD}) and writer independent (\textit{WI}) modules. }
\label{ablation study1}
\resizebox{.48\textwidth}{!}{%
\begin{tabular}{cccllll}
\hline
\multicolumn{1}{l}{}       & \multicolumn{2}{c}{IAM} & \multicolumn{2}{c}{CVL}                                      & \multicolumn{2}{c}{CERUG-EN}                                 \\ \cline{2-7} 
Configuration     & Top 1            & Top 5           & \multicolumn{1}{c}{Top 1} & \multicolumn{1}{c}{Top 5} & \multicolumn{1}{c}{Top 1} & \multicolumn{1}{c}{Top 5} \\ \hline
Baseline (\emph{\textit{WD}})                    & 92.07                             & 96.43     & 89.76           & 95.46                             & 96.05                             & 99.56                             \\
\textit{WI+\textit{WD}+Max}         & 93.13                             & 96.51     & 90.58           & 96.25                               & 97.08                             & 100                               \\
\textit{WI+\textit{WD}+Add}            & 93.43                             & 97.35     & 91.12           & 96.79                               & 97.18                             & 100                               \\
\textit{WI+\textit{WD}+Concat}      & 93.75                             & 97.46  & 91.86           & 97.11                                & 97.56                             & 100                               \\ \hline
\end{tabular}%
}
\end{table}

\subsection{Impact of the dual stream network with attention}
We now investigate the impact of incorporating an attention module on the overall performance of the system, as discussed in section \ref{attn}. Explicitly, two configurations of the attention module presented in Fig \ref{attention_config} (a) and (b) are analyzed using the \textit{WI+WD+Concat} framework.

Table \ref{combined_table} summarizes the results obtained from these configurations. Based on the entries, it may be inferred that integrating the attention mechanism impacts the overall performance of the system. Upon comparing the baseline with the attention configuration in Fig \ref{attention_config} (a), the identification precision shows notable improvements with Top 1 accuracy increasing from 92.07 \%, 89.76 \%, 96.05 \% to 93.92 \%, 92.62 \% and 97.71 \%,  for the IAM, CVL, and CERUG-EN datasets, respectively. Likewise, the configuration in Fig \ref{attention_config} (b) results in best  accuracy of  94.43 \%, 93.15 \% and 97.86 \%. The results pertain to a multi-head attention setup with 2 attention heads and a dimensionality of 64 for the query, key, and value vectors. This configuration is chosen based on empirical observation that increasing the number of heads does not improve overall system performance. Additionally, increasing the dimension of the query vectors tends to reduce performance, suggesting that a compact representation is more effective.

\begin{table}[t]
\centering
\caption{Impact of the attention framework configurations of Fig 3 on the average identification rate (\%) over the baseline \textit{WD} model (with no attention).}
\label{combined_table}
\setlength{\tabcolsep}{3pt} 
\begin{tabular}{@{}ccccccc@{}}
\toprule
& \multicolumn{2}{c}{{IAM}} & \multicolumn{2}{c}{{CVL}} & \multicolumn{2}{c}{{CERUG}} \\
\cmidrule{2-3} \cmidrule{4-5} \cmidrule{6-7}
Attention configuration & Top 1 & Top 5 & Top 1 & Top 5 & Top 1 & Top 5 \\
\midrule
\makecell{No attention}                    & 92.07                             & 96.43     & 89.76           & 95.46                             & 96.05                             & 99.56                             \\
\makecell{Fig 3 (a)} & 93.92 & 97.65 & 92.62 & 97.79 & 97.71 & 100 \\
\makecell{Fig 3 (b)} & 94.43 & 98.26 & 93.15 & 98.38 & 97.86 & 100 \\
\bottomrule
\end{tabular}
\end{table}

\begin{table}[t]
\centering
\caption{Performance of writer identification system with \emph{\textit{WD}} module trained on word and fragments.}
\label{wrd_frag}
\begin{tabular*}{.48\textwidth}{@{\extracolsep\fill}ccccc}
\toprule%
& \multicolumn{2}{@{}c@{}}{{Word}} & \multicolumn{2}{@{}c@{}}{{Fragment}}   \\\cmidrule{2-3}\cmidrule{4-5}%
{Database} & Top 1 & Top 5 & Top 1 & Top 5  \\
\midrule
CVL & 87.76 & 94.60 & 89.76 & 95.46 \\
IAM & 86.08 & 92.94 & 92.07 & 96.43 \\
CERUG-EN & 65.76& 87.52  & 96.05 & 99.56 \\ 
\bottomrule
\end{tabular*}
\end{table}
\subsection{Efficacy of fragment driven training of the dual network}
\begin{figure}[t]
\centering
\includegraphics[width=.48\textwidth,keepaspectratio]{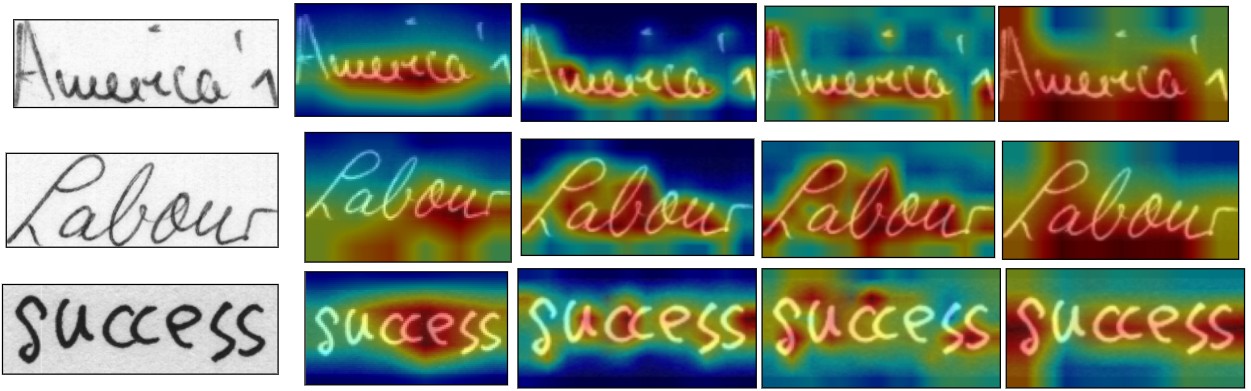}
\caption{Visualization of heatmap activation across different network configurations. Rows 1–3 correspond to a word sample from the the IAM, CVL, and CERUG-EN datasets, respectively. Column 2 shows the heatmap produced by the \textit{WD} module trained on complete word images, while columns 3–5 display heatmaps obtained from networks trained  with the (i) the \textit{WD} module, (ii) combined (\textit{\textit{WD}+WI+Concat}) module without attention and (iii) combined (\textit{\textit{WD}+WI+Concat}) module with the attention configuration of Fig \ref{attention_config}(b), respectively.}
\label{fragment_hmap}
\end{figure}

Moving further, we would like to investigate how the baseline CNN network (\emph{\textit{WD}} module) would perform when trained on whole word images in place of its constituent fragments. It is interesting to see that the fragment level training  of the network outperforms over that of  whole word images (refer Table \ref{wrd_frag}). This trend  may be attributed to the fact that   convolution networks trained on word images make decisions based on the holistic features, which may not fully capture all the relevant details of the writing styles. Contrast to that, our fragment-based training approach treats each image as a collection of local patches, thereby possibly allowing the network to learn more relevant features. This results in an enhanced representation and  in turn explains the effectiveness of the proposed identification process.\\
\indent For completeness, we also present the influence of CNNs trained on both word images and image fragments using heat map visualizations, corresponding to a sample from the IAM, CVL and CERUG-EN dataset (Figure~\ref{fragment_hmap}). These heat maps highlight the regions of the input emphasized by the network. Specifically, column 2 shows the activation map from the  baseline \emph{\textit{WD}} module trained on complete word images, while columns (3–5) display activation maps generated by models trained on image fragments under  (i) the \emph{\textit{WD}} module, (ii) combined (\textit{\textit{WD}+WI+Concat}) module without attention and (iii) combined (\textit{\textit{WD}+WI+Concat}) module with the attention configuration of Fig \ref{attention_config}(b), respectively. Red regions indicate higher
 attention focus by the network, while blue regions represent
 lower attention. Note that all heatmaps are post-processed
 using Gaussian smoothing to reduce blockiness introduced by
fragment-wise reconstruction and to enhance visual coherence. \\  \indent The model trained on complete word images (column 2) produces focused activations only on selective parts of the word, while some regions get unnoticed. Contrast to it, the attention-based configuration (column 5) yields well focused activations, that closely align with handwriting strokes and highlight discriminative regions in the fragments, thereby demonstrating its ability to capture writer features.  Further, the higher attention in the heatmaps in Column 5 aligns
 with the improved performance of the proposed architecture of Figure \ref{cnn_model} (a) , when  trained \textbf{solely} by using the fragments of the word.

\subsection{Performance comparison with prior end-to-end  architectures} \label{compar}
In this sub-section, we compare the performance of our proposed end-to-end network with  recent deep-learning networks for writer identification of handwritten word image. The result of this analysis is presented in Table \ref{prior_wrk}. Our proposed method trained using fragments in a dual path CNN framework performs on par with state-of-the-art methods that use both word data together with their constituents fragments for learning.   It is interesting to note that our method also outperforms the  work \cite{Zhang:2022}, that employs the Swin transformer  implemented using the position encoding information. However,  considering that our method is text independent , it  does not necessitate a strict positioning of the fragments  within a word. This is an important  aspect that we believe is  overlooked in the  residual swin transformer,  leading to a slightly lower performance.

\begin{table}[t]
\centering
\caption{Performance comparison with prior end-to-end  architectures}
\label{prior_wrk}
\setlength{\tabcolsep}{3pt} 
\begin{tabular*}{0.48\textwidth}{@{\extracolsep\fill}ccccccc}
\toprule
& \multicolumn{2}{c}{{IAM}} & \multicolumn{2}{c}{{CVL}} & \multicolumn{2}{c}{{CERUG}} \\
\cmidrule{2-3} \cmidrule{4-5} \cmidrule{6-7}
Method & Top 1 & Top 5 & Top 1 & Top 5 & Top 1 & Top 5 \\
\midrule
Deep-Adapt. \cite{He:2019}       & 69.5 & 86.1 & 79.1 & 93.7 & - & - \\
ResNet18+HTD \cite{Javidi:2020}          &76.9 & 91.6 & 85.1 & 95.6 & 70.1 & 91.8 \\
WordImgNet \cite{He:2020}        & 81.8 & 94.1 & 88.6 & 96.8 & 77.3 & 96.4 \\
ResNet18 \cite{He:2016}          & 83.2 & 94.3 & 88.5 & 96.7 & 70.6 & 94.0 \\
Patch-net \cite{Srivastava:2022}          & 80.2 & 93.5 & 86.1& 96.2 & 77.1 & 96.5 \\
SA-Net \cite{Srivastava:2022}    & 83.4 & 94.6 & 90.7 & 97.4 & 82.2 & 97.1 \\
MSRF-Net\cite{Srivastava:2022}    & 84.6 & 95.0 & 91.4 & 97.6 &79.6 & 96.8 \\
FragNet-64 \cite{He:2020}        & 85.1 & 95.0 & 90.2 & 97.5 & 77.5 & 95.6 \\
Vert. FGRR \cite{He:2021}        & 85.9 & 95.2 & 92.6 & 97.9 & 82.6 & 95.8 \\
Horiz. FGRR \cite{He:2021}       & 86.1 & 95.0 & 92.4 & 97.8 & 83.2 & 96.2 \\
Multistage CNN with M-DCN \cite{Okawa:2025}  & 86.7 & 95.2 & 92.1 & 97.9 &86.8 & 97.2 \\
Residual Swin Transformer  \cite{Zhang:2022}           & 90.7 & 96.6 & 92.7 & 97.9 & 87.6  & 98.7 \\
{Proposed }         & {94.4} & {98.3} & {93.2} & {98.4} & {97.9} & {100} \\
\bottomrule
\end{tabular*}
\end{table}

\section{Conclusion} \label{sec8}
This study explored a  dual-stream convolution-based End-to-end network designed for offline text-independent writer identification systems focusing on word images. Our proposed network combines writer-specific local features with writer-independent global features to produce a strong representation of writer characteristics.  Furthermore, the effect of integrating an attention mechanism on the overall performance of the system was investigated. Experimental results indicate the superiority of our proposed network in scenarios with limited samples when compared to prior networks trained on word images in terms of  accuracy.






\bibliographystyle{IEEEtran}
\bibliography{bibtex.bib}

\end{document}